\documentclass{article}
\usepackage{ijcai21}

\usepackage{times}
\usepackage{soul}
\usepackage{url}
\usepackage[hidelinks]{hyperref}
\usepackage[utf8]{inputenc}
\usepackage[small]{caption}
\usepackage{graphicx}
\usepackage{amsmath}
\usepackage{amsthm}
\usepackage{booktabs}
\usepackage{algorithm}
\usepackage{algorithmic}
\usepackage{booktabs}
\usepackage{xcolor}
\title{Bandit Modeling of Map Selection in Counter-Strike: Global Offensive}

\author{
Guido Petri$^{*1}$\and
Michael H. Stanley$^{*1}$\and
Alec B. Hon$^{*1}$\and 
Alexander Dong$^{*1}$\and \\
Peter Xenopoulos$^{1}$\and
Cláudio Silva$^{1}$
\affiliations
$^1$New York University\\
\emails
\{gp1655, mhs592, abh466, awd275, xenopoulos, csilva\}@nyu.edu

}

\begin{document}

\maketitle

\begin{abstract}
  Many esports use a pick and ban process to define the parameters of a match before it starts. In Counter-Strike: Global Offensive (CSGO) matches, two teams first pick and ban maps, or virtual worlds, to play. Teams typically ban and pick maps based on a variety of factors, such as banning maps which they do not practice, or choosing maps based on the team's recent performance. We introduce a contextual bandit framework to tackle the problem of map selection in CSGO and to investigate teams' pick and ban decision-making. Using a data set of over 3,500 CSGO matches and over 25,000 map selection decisions, we consider different framings for the problem, different contexts, and different reward metrics. We find that teams have suboptimal map choice policies with respect to both picking and banning. We also define an approach for rewarding bans, which has not been explored in the bandit setting, and find that incorporating ban rewards improves model performance. Finally, we determine that usage of our model could improve teams' predicted map win probability by up to 11\% and raise overall match win probabilities by 19.8\% for evenly-matched teams.
\end{abstract}
\let\thefootnote\relax\footnotetext{$^*$ Equal contribution.}
\section{Introduction}
As data acquisition methods become more pervasive, sports analytics has received increased interest in contemporary sports, like soccer, basketball and baseball~\cite{DBLP:journals/bigdata/AssuncaoP18}. One common application in sports analytics is valuing player actions and decision-making. For example, Decroos~et~al.~introduce a framework to value soccer players according to how their actions change their team's chance of scoring~\cite{DBLP:conf/kdd/DecroosBHD19}.

Esports, also known as professional video gaming, is one of the fastest growing sports markets in the world. Yet esports has attracted little sports analytics interest. Most analytical work in esports covers massively online battle arena (MOBA) games, such as League of Legends or Defense of the Ancients 2 (``DOTA2"). Accordingly, there exists a dearth of work on Counter-Strike: Global Offensive (CSGO), one of the oldest yet most popular esports. A picking and banning process is a common process in many esports, where some entities are banned from being played in a particular game. For example, in League of Legends, teams ban a set of characters, and their players each pick a character to play before the game starts. In CSGO, teams typically perform a map selection process where each team takes turns picking and banning maps to play. However, map selection routines are often not based on analytics and data, but rather on players' inclinations at selection time.

Contextual bandits are statistical models that take a context $x$ and return a probability distribution over possible actions $a$, with the objective of maximizing the reward $r$ returned by the action taken. In this paper, we apply a contextual bandit framework to the domain of map selection in CSGO. We use a novel data set of over 25,000 map pick and ban decisions from over 3,500 professional CSGO matches to train three different bandit framings to the problem. We find that teams' choices in the map selection process are suboptimal and do not yield the highest expected win probability.

The paper is structured accordingly. In section~\ref{section:RelatedWork}, we review relevant esports and contextual bandit works. In section~\ref{section:CSMapSelection}, we cover CSGO's map selection process. In section~\ref{section:Modeling}, we introduce our contextual bandit model. In section~\ref{section:Experiments}, we describe our dataset and our evaluation methodology. Section~\ref{section:Results} contains our results. We discuss the benefits of our model, the choices of evaluation metrics and suggest possible areas of future work in section~\ref{section:Discussion} and conclude the paper in section~\ref{section:Conclusion}.


\begin{figure*}
    \center{\includegraphics[width=\linewidth]
    {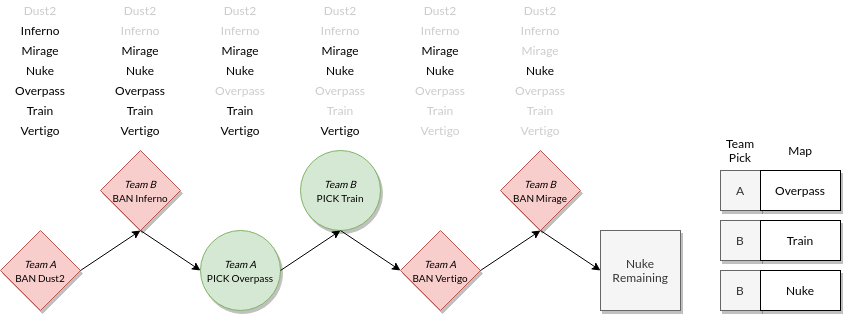}}
    \caption{\label{fig:mappicking} Example map selection process for a best-of-three match. The available map pool is shown above each pick/ban decision. The first team, usually decided by tournament rules, bans a map. The second team then does the same. The two teams then both pick a map, and then both ban a map. In total, there are six decisions, four of which are bans, and two are picks.}
\end{figure*}

\section{Related Work} \label{section:RelatedWork}

Reinforcement learning (RL) techniques are increasingly being applied to sports analytics problems. Liu~et~al.~first used RL in sports to estimate an action-value Q function from millions of NHL plays~\cite{DBLP:conf/ijcai/LiuS18}. They used the learned Q function to value players based on the aggregate value of their actions. Liu~et~al.~also apply mimic learning to make their models more interpretable~\cite{DBLP:conf/pkdd/LiuZS18a}. Sun~et~al.~extend this work by considering a linear model tree~\cite{DBLP:conf/kdd/SunDSL20}. While the previous works heavily focused on ice hockey, Liu~et~al.~also learn an action-value Q function for soccer~\cite{DBLP:journals/datamine/LiuLSK20}. Despite the heavy use of other RL approaches such as Q-learning, contextual bandits have not been as heavily utilized in sports analytics.

This paper applies contextual bandits to the multi-arm map selection process in esports matches for the game CSGO. Contextual bandits are a simplified case of reinforcement learning. In reinforcement learning, an action is chosen based on the context (or state) and a reward is observed, and this process is repeated for many rounds. Rewards are not observed for actions not chosen. In the contextual bandit case, the contexts of different rounds are independent. \cite{tewari:context_bandit} provides a thorough review of contextual bandits, tracing the concept back to \cite{woodroofe:context_bandit} and the term back to \cite{langford:context_bandit}. Many approaches have been explored for learning policies in the contextual bandit setting. \cite{williams:rl} introduced gradient approaches in the reinforcement learning setting, and \cite{sutton_bartow:rl} applied the approach to the specific case of contextual bandits. Comparing proposed policies often requires off-policy evaluation: estimating the value of a policy from data that was generated by a different policy (the ``logging policy"). This paper utilizes two off-policy evaluation approaches: the self-normalized importance-weighted estimator \cite{swaminathan:sn-iw} and the direct method of regression imputation \cite{Dud_k_2014}. To our knowledge, ban actions have never been modeled in the bandit setting. 

Esports have mostly attracted sports analytics interest in the form of win prediction and player valuation. Numerous efforts have been made to predict win probabilities in popular esports games such as CSGO and DOTA2. \cite{yang:match_pred} and \cite{hodge:match_pred} first use logistic regression and ensemble methods to predict win probabilities in DOTA2, a popular MOBA game. \cite{makarov:csgo} first predicted CSGO win probabilities using logistic regression, however their data only included less than 200 games. \cite{xenopoulos:csgo} expand on previous CSGO work by introducing a data parser and an XGBoost based win probability model for CSGO. They also value players based on how their actions change their team's chance of winning a round. \cite{bednarek:csgo} value players by clustering death locations. 

Map selection is a process largely unique to CSGO and has not been well studied, but is loosely related to another esports process unique to MOBA games: hero selection. In DOTA2, for example, players from opposing teams alternate choosing from over one hundred heroes, with full knowledge of previous hero selections. \cite{yang:match_pred} and \cite{song:dota} use the selected heroes as features to predict win probability, but do not recommend hero selections or explicitly model the selection process.  More relevant is the hero selection recommendation engine of \cite{conley:dota}, which uses logistic regression and K-nearest neighbors to rank available heroes based on estimated win probability; they do not, however, consider historical team or player context.

\section{Counter-Strike Map Selection}\label{section:CSMapSelection}
Counter-Strike is a popular esport that first came out in 2000, and CSGO is the latest version. The game mechanics have largely stayed the same since the first version of the game. Before a CSGO match starts, two teams go through the map selection process to decide which maps the teams will play for that match. A map is a virtual world where CSGO takes place. Typically, matches are structured as a best-of-three, meaning the team that wins two out of three maps wins the match. A team wins a map by winning rounds, which are won by completing objectives.


The collection of available maps in the map selection process is called the \textit{map pool}. Typically, there are seven maps to choose from in the map pool. Although the maps rarely change, a new map may be introduced and replace an existing map. Our data contains map selections using the following map pool: \texttt{dust2}, \texttt{train}, \texttt{mirage}, \texttt{inferno}, \texttt{nuke}, \texttt{overpass}, \texttt{vertigo}. The map selection process is exemplified in Figure~\ref{fig:mappicking}. First, team A \textit{bans} a map. This means that the teams will not play the map in the match. The team that goes first in the map selection process is usually higher seeded, or determined through tournament rules. Next, team B will ban a map. The teams then will each \textit{pick} a map that they wish to play in the match. Next, team A will ban one more map. At this point, team B will ban one of the two remaining maps, and the map not yet picked or banned is called the \textit{decider}. 


Professional teams may sometimes have what is referred to as a \textit{permaban} -- a map that they will always ban with their first ban. For example, some teams may choose to ban the same map in over 75\% of their matches. From interviews with four CSGO teams ranked in the top 30, two of which are in the top 10, teams choose their maps from a variety of factors. Some teams try to choose maps they have a historically high win percentage, or maps where their opponents have low win percentages. Other teams may also choose maps purely based on how their recent practice matches performances. 


\section{Bandit Model for CSGO Map Selection} \label{section:Modeling}

In order to model the map selection process, we elected to use a \textit{k}-armed contextual bandit. This was a clear choice: the actions taken by teams only yield a single shared reality, where we cannot observe the counterfactual of different choices. The bandit model enables us to approximate the counterfactual reality and frame this problem as a counterfactual learning problem.

In particular, we used the context from teams' previous matches, as well as information available at the time of selection, such as which maps were still in the selection pool. There are two kinds of actions: picks and bans, which must be manipulated differently. The reward is the map being won by the choosing team or not, as well as more granular version of this in which we include margin of victory.

\subsection{Context and Actions}

Our initial choice for the context given a particular round $t$ in the map-picking process was a one-hot encoding for the available maps in that particular round, such that the bandit would learn to not pick the map if it was not available. To give the bandit more information about the teams that were deciding for that particular match, we implemented two historical win percentages, the first being the team's historical match win percentage, and the second being the team's historical map win percentage for each map. The first percentage is utilized to indicate team strength compared to other teams, and the second the team's overall ability to play well on each map. We applied Laplace smoothing to the initial percentages for numerical stability, using the formula

\begin{equation}
    \text{Win\%} = \dfrac{\text{Wins} + 5}{\text{Matches} + 10}.
\end{equation}

Both win percentages were stored in the context vector for both the deciding team and the opponent team alongside the available maps. For both picks and bans, the given \textit{context} is the same as described above, and the corresponding \textit{action} would be the map picked or banned by the deciding team.
\subsection{Rewards}
\subsubsection{Picks}

Due to the nature of the map-picking process, where the decider is a forced pick, we chose to remove the rewards from all final map picks, as it would not make sense to reward either team for a forced choice. As a result, only the first two picks from each map selection process were given a reward. Rewards for map-picking were implemented with two different methods. Our first method utilized a simple 0-1 reward (``0/1"), where if the deciding team won on the map they had picked, they would be rewarded with an overall reward of $1$ for that action, or $0$ otherwise. Our second method rewarded the deciding team based on the margin of rounds won (``MoR") in the best-of-30 rounds on the decided map. The reward function for deciding team $i$ and an opponent team $j$ is given below:

\begin{equation}
    R_{i,j}= \frac{\text{Rounds won by  $i$} - \text{Rounds won by $j$}}{\text{Total number of Rounds on map}} 
\end{equation}

The round proportion rewards were implemented as a more granular method to compare team performance on each map.

\subsubsection{Bans}

Since there is no data on how any deciding team would perform on a banned map, we chose to reward bans based on the deciding team's overall performance in the match, where if the deciding team won the match, they would be rewarded for choosing to not play on the banned map with an overall reward of $1$, or, if they lost, a reward of $-1$. In addition, we implemented a exponentially decreasing reward over the ban process, where earlier bans would have higher rewards. Later map picks have fewer available choices: restricting the action space means a team may be forced to make a choice they do not want, and so we de-emphasize the later choices. The ban reward function for team $i$ playing in match $t$ is given below:
\begin{equation}
    R_{i,t}(n) = 
    \begin{cases}
         1 \cdot \frac{1}{2^n} & \text{if team $i$ won match $t$} \\
         -1 \cdot \frac{1}{2^n}& \text{if team $i$ lost match $t$} \\
  \end{cases}
\end{equation}
where $n$ is the $n$th ban in the map picking process. In our case, $n \in \{1,2,3,4\}$, as there are always four bans in the map picking process for CSGO. 

\subsection{Policy Gradient Learning}

The most straightforward way to train a bandit is via policy gradient learning \cite{sutton_bartow:rl}. For our policy class, we use a multinomial logistic regression parameterized by weights $\theta$ and an action-context mapping function $\phi(x, a)$, with the softmax function to transform the affinity of the bandit for each action into a probability:

\begin{equation}
\pi(a|x) = \dfrac{\exp(\theta^{T} \phi(x, a))}{\sum_{i=1}^k \exp(\theta^{T} \phi(x, i))}
\end{equation}

The policy gradient approach trains the model via SGD \cite{sutton_bartow:rl}, enabling both online and episodic learning. In particular, the optimization maximizes the expected reward for the bandit, using the update function

\begin{table*}[]
\centering
\begin{tabular}{@{}lrrrr@{}}
\toprule
                       & \multicolumn{1}{c}{Picks (0/1)} & \multicolumn{1}{c}{Picks (MoR)} & \multicolumn{1}{c}{Bans (0/1)} & 
                       \multicolumn{1}{c}{Bans (MoR)} \\ \midrule
Uniform policy (split) & 0.568/0.541                    & 0.568/0.541                          & -0.018/-0.003                    & -0.018/-0.003                         \\
Logging policy         & 0.549/0.549                       & 0.549/0.549                          & -0.014/-0.014                 & -0.014/-0.014                         \\
SplitBandit            & 0.587/0.554                      & \textbf{0.659/0.528}                 & -0.016/0.004                       & -0.016/0.004                         \\
ComboBandit             & \textbf{0.640/0.528}              & 0.613/0.573                          & \textbf{0.021/0.003}               & \textbf{0.036/-0.015}                 \\
EpisodicBandit         & 0.568/0.551                       & 0.561/0.547                          & 0.013/0.006                    & 0.013/0.006                           \\ \bottomrule
\end{tabular}
\caption{Expected reward for each policy type under four different evaluations. The best policy parameters were found via grid search and the policy was optimized with policy gradient. Both the SN-IW (left) and DM (right) evaluation methods are presented, except for Logging policy where the on-policy value is presented. Every model tested outperforms or matches the baseline uniform policy, with the best overall model being the bandit trained on both picks and bans. Comparisons between the uniform and logging policy indicate teams choose their bans well, but their picks poorly.}
\label{table:mainresults}
\end{table*}

\begin{equation}
\theta_{t+1} \leftarrow \theta + \eta R_t(A_t) \nabla_{\theta} \log \pi_{\theta_t}(A_t|X_t)
\end{equation}

with $\pi$ defined above and the gradient

\begin{equation}
\resizebox{.91\linewidth}{!}{$\nabla_{\theta} \log \pi(a|x) = \phi(x,a) - \dfrac{\sum_{i=1}^k \phi(x, i)  \exp(\theta^T \phi(x, i))}{\sum_{i=1}^k \exp(\theta^T \phi(x, i))}.$}
\end{equation}

In the context of picks, we can use online learning to iteratively update the parameters $\theta$. For bans, however, we do not observe a reward at the time the choices are made; as a result, we used episodic learning, where an episode is an entire match.

\section{Experiments}\label{section:Experiments}
\subsection{Data}
We obtained our data from HLTV.org, a popular CSGO fan site. The site contains scores, statistics and map selections for most professional matches. We use matches from April 2020 to March 2021. In total, this consisted 628 teams that played a total of 6283 matches, summing to 13154 games. We only consider best-of-three matches, which are by far the most popular match format. We focus on the games where the most common set of seven maps is selected from the map pool of \texttt{dust2}, \texttt{inferno}, \texttt{mirage}, \texttt{nuke}, \texttt{overpass}, \texttt{train}, \texttt{vertigo}. In addition, we also remove teams such that in the final dataset, each team has played at least 25 games, or approximately 10 matches, with another team in the dataset. This leaves us with 165 teams, playing a total of 3595 matches, summing to 8753 games. The resulting dataset was split into an 80-20 train-test split by matches for bandit learning and evaluation.

\subsection{Evaluation}
We use two typical off-policy evaluation methods, the \textit{direct method}  (``DM") \cite{Dud_k_2014} and the self-normalized importance-weighted estimator (``SN-IW") \cite{swaminathan:sn-iw}. We also present the mean reward observed as a baseline. 

The goal of the direct method is to estimate the reward function $r(x, a)$ that returns the reward for any given action $a$ for the context $x$. We estimate the reward function by using an importance-weighted ridge regression for each action. We use the self-normalized importance-weighted estimator with no modifications.

Value estimates are presented for four different reward and model training settings:
\begin{itemize}
    \item Picks(0/1): Expected pick reward for models trained with 0/1 rewards
    \item Picks(MoR): Expected pick reward for models trained with MoR rewards
    \item Bans(0/1): Expected ban reward for a models trained with 0/1 rewards
    \item Bans(MoR): Expected ban reward for models trained with MoR rewards
\end{itemize}

\subsection{Variety of Policies}

We experimented with three different varieties of contextual bandits: \texttt{SplitBandit}, \texttt{ComboBandit}, and \texttt{EpisodicBandit}.  

\texttt{SplitBandit} is composed of two individual, simple contextual bandits, each with a $\theta$ parameter size of $(\texttt{n\_features} \cdot \texttt{n\_arms})$. The first contextual bandit is trained on the picks via online learning. The second contextual bandit is trained on the bans in an episodic fashion.

\texttt{ComboBandit} is a single model also trained on the picks via online learning and on the bans via episodic learning with a $\theta$ parameter size of $(\texttt{n\_features} \cdot \texttt{n\_arms})$. \texttt{ComboBandit} learns a single set of parameters that define a policy for both picks and bans. The ban policy is derived from the pick policy:
\begin{equation}
    \pi_B(a|X) = \frac{1-\pi_P(a|X)}{\sum_{\alpha \in A}1-\pi_P(\alpha|X)}
\end{equation}
for pick policy $\pi_P$ and ban policy $\pi_B$ over actions $A$ and context $X$.

\texttt{EpisodicBandit} is similarly a single model, but it is trained on both the picks and bans simultaneously via episodic learning with a $\theta$ parameter size of $(2 \cdot \texttt{n\_features} \cdot \texttt{n\_arms})$. We expected this model to perform similarly to \texttt{SplitBandit}, since its gradient estimates are better estimates than the estimates derived from individual datapoints, offsetting the quicker adaptability of the online gradient calculation with less noise.

\section{Results} \label{section:Results}

Our main results are summarized in table~\ref{table:mainresults}. Considering the self-normalized estimator, the best model for picks was \texttt{SplitBandit} trained on proportional rewards, while the best model for bans was \texttt{ComboBandit} trained on proportional rewards. The uniform policy performs better than the logging policy for the picks in our dataset but worse for bans, which indicates teams' picks might be overconfident, whereas their bans are chosen more carefully.

\begin{figure}[t]
    {\includegraphics[width=9cm]
    {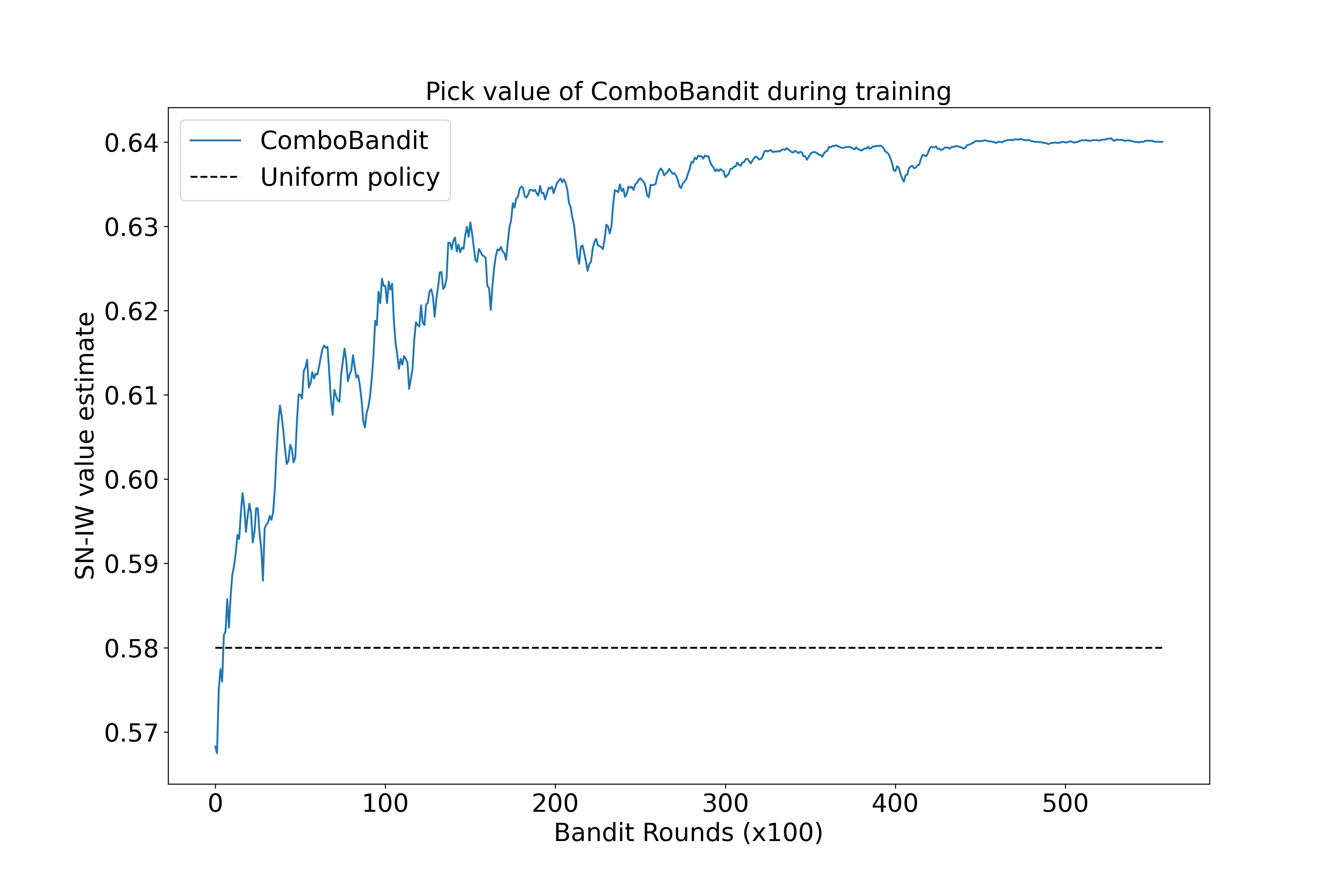}}
    \caption{\label{fig:value_over_time} Picks(0/1) value on the test set for \texttt{ComboBandit} and Uniform policy, evaluated every 100 rounds over 3 epochs of training. The bandit quickly surpasses the uniform policy's performance and plateaus around an expected reward value of approximately $0.64$.}
\end{figure}

\texttt{ComboBandit} substantially outperforms all other policies. We believe this is due to its training including the additional data from both picks and bans instead of selecting only one of the two categories for training a given parameter. This yields a better optimization through better gradient estimates. \texttt{EpisodicBandit} is trained on both picks and bans, but its parameters do not depend on both subsets of data, which does not provide that optimization advantage. The learning curve in Figure \ref{fig:value_over_time} shows that ComboBandit surpasses the uniform policy benchmark after only a few training rounds, continuing to improve over 3 epochs of training.

\begin{figure}[t]
    {\includegraphics[width=9cm]
    {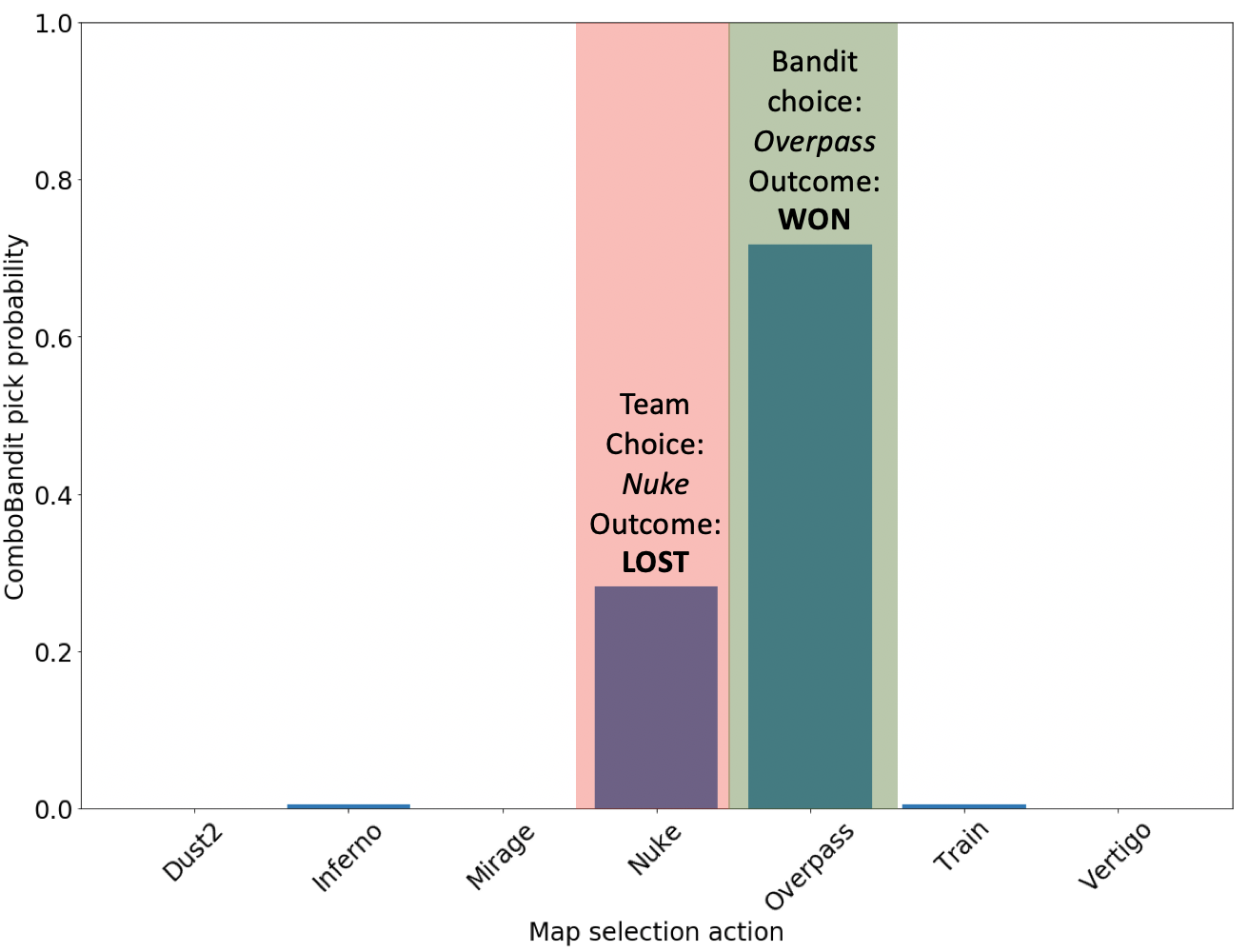}}
    \caption{\label{fig:policy_comparison} The best model's probability distribution for pick 4 in a match between \textit{TIGER} and \textit{Beyond}. \textit{TIGER}, the deciding team, chose \texttt{Nuke} and lost the map, later going on to win map \texttt{Overpass}, which was \texttt{ComboBandit}'s suggestion.}
\end{figure}

Figure~\ref{fig:policy_comparison} shows an example of \texttt{ComboBandit}'s policy. In this match, team \textit{TIGER} chose to play on the map \texttt{Nuke}, which they later lost. \texttt{ComboBandit} suggested instead to play on \texttt{Overpass}, with 71\% probability. In the same match, \texttt{Overpass} was chosen as the decider and \textit{TIGER} won that map, indicating that the bandit model's policy distribution was more valuable than the team's intuition on map choice.


\section{Discussion} \label{section:Discussion}

The results indicate that teams using our chosen policy instead of their traditional map-picking process can increase their expected win probability by 9 to 11 percentage points, depending on the policy used. This is a substantial advantage for a best-of-3 match, since the model could confer that added win probability to all three map choices. The ban choice can be improved as well by using our model. The logging policy yields an expected reward of approximately $-0.014$, which indicates that bans have a slight negative effect on match win probability. However, our best model's expected reward for bans is $0.036$, thus increasing match win probability by approximately 5\% after a ban choice. For two teams that are evenly matched, using our bandit for both pick and ban decisions translates to the team that uses the model having an expected overall match win probability of 69.8\% instead of 50\%, a substantial advantage for a team.

The choice of evaluation metric is particularly important in examining the results. Using the direct method instead of the self-normalized estimator, we reach drastically different conclusions about which model to use, with the best overall model being \texttt{EpisodicBandit}. In our experiments, we used ridge regressions for our regression imputation. This is clearly a suboptimal model for this estimation, since the context features of win probabilities are bounded: there is a non-linear relationship between the context and the rewards. This is a big limitation of our experiments: we instead relied on the importance-weighted estimator, which is known to be imprecise in estimating policies far from the logging policy.

Future work in this area will be concentrated on examining better choices for evaluation metrics, as well as expanding the contextual features further by adding, for example, player turnover, team-based Elo metrics or rankings, or examining recent performances, such as win percentage in the last 10 matches. The rewards can also be expanded by using not only margin of rounds won per map, but also the margin of players alive per map at the end of a round. Additionally, different framings for the bandit can be considered, such as creating a ranking of which maps are best to choose instead of the model selecting a single map for the user. 


\section{Conclusion} \label{section:Conclusion}

We modeled the map selection process in Counter-Strike: Global Offensive as a bandit, framing the problem in several different ways. Our key contributions are (1) the introduction of bandits and simple reinforcement learning models to esports and CSGO in particular, and (2) novel ways of implementing negative choices in bandits, for which we explicitly choose not to observe their rewards. We find that our model shows that teams are making sub-optimal map selections. 


\section*{Acknowledgments}
This work was partially supported by: NSF awards CNS-1229185, CCF-1533564, CNS-1544753, CNS-1730396, CNS-1828576, CNS-1626098. We additionally thank David Rosenberg.

\appendix

\bibliographystyle{named}
\bibliography{ijcai21}

\begin{thebibliography}{}

\bibitem[\protect\citeauthoryear{Assun{\c{c}}{\~{a}}o and
  Pelechrinis}{2018}]{DBLP:journals/bigdata/AssuncaoP18}
Renato~M. Assun{\c{c}}{\~{a}}o and Konstantinos Pelechrinis.
\newblock Sports analytics in the era of big data: Moving toward the next
  frontier.
\newblock {\em Big Data}, 6(4):237--238, 2018.

\bibitem[\protect\citeauthoryear{Bednárek \bgroup \em et al.\egroup
  }{2017}]{bednarek:csgo}
David Bednárek, Martin Kruliš, Jakub, Yaghob, and Filip Zavoral.
\newblock Player performance evaluation in team-based first-person shooter
  esport.
\newblock {\em International Conference on Data Management Technologies and
  Applications}, pages 154--175, July 2017.

\bibitem[\protect\citeauthoryear{Conly and Perry}{2017}]{conley:dota}
Kevin Conly and Daniel Perry.
\newblock How does he saw me? {A} recommendation engine for picking heroes in
  {Dota 2}.
\newblock Unpublished course project, 2017.

\bibitem[\protect\citeauthoryear{Decroos \bgroup \em et al.\egroup
  }{2019}]{DBLP:conf/kdd/DecroosBHD19}
Tom Decroos, Lotte Bransen, Jan~Van Haaren, and Jesse Davis.
\newblock Actions speak louder than goals: Valuing player actions in soccer.
\newblock In Ankur Teredesai, Vipin Kumar, Ying Li, R{\'{o}}mer Rosales,
  Evimaria Terzi, and George Karypis, editors, {\em Proceedings of the 25th
  {ACM} {SIGKDD} International Conference on Knowledge Discovery {\&} Data
  Mining, {KDD} 2019, Anchorage, AK, USA, August 4-8, 2019}, pages 1851--1861.
  {ACM}, 2019.

\bibitem[\protect\citeauthoryear{Dudík \bgroup \em et al.\egroup
  }{2014}]{Dud_k_2014}
Miroslav Dudík, Dumitru Erhan, John Langford, and Lihong Li.
\newblock Doubly robust policy evaluation and optimization.
\newblock {\em Statistical Science}, 29(4), Nov 2014.

\bibitem[\protect\citeauthoryear{Hodge \bgroup \em et al.\egroup
  }{2019}]{hodge:match_pred}
Victoria~J. Hodge, Sam Devlin, Nick Sephton, Florian Block, and Peter~I.
  Cowling.
\newblock Win prediction in multi-player esports: Live professional match
  prediction.
\newblock {\em IEEE Transactions on Games}, 2019.

\bibitem[\protect\citeauthoryear{Langford and
  Zhang}{2008}]{langford:context_bandit}
John Langford and Tong Zhang.
\newblock The epoch-greedy algorithm for multi-armed bandits with side
  information.
\newblock {\em In Advances in neural information processing systems}, pages
  817--824, 2008.

\bibitem[\protect\citeauthoryear{Liu and
  Schulte}{2018}]{DBLP:conf/ijcai/LiuS18}
Guiliang Liu and Oliver Schulte.
\newblock Deep reinforcement learning in ice hockey for context-aware player
  evaluation.
\newblock In J{\'{e}}r{\^{o}}me Lang, editor, {\em Proceedings of the
  Twenty-Seventh International Joint Conference on Artificial Intelligence,
  {IJCAI} 2018, July 13-19, 2018, Stockholm, Sweden}, pages 3442--3448.
  ijcai.org, 2018.

\bibitem[\protect\citeauthoryear{Liu \bgroup \em et al.\egroup
  }{2018}]{DBLP:conf/pkdd/LiuZS18a}
Guiliang Liu, Wang Zhu, and Oliver Schulte.
\newblock Interpreting deep sports analytics: Valuing actions and players in
  the {NHL}.
\newblock In {\em Proceedings of the 5th Workshop on Machine Learning and Data
  Mining for Sports Analytics co-located with 2018 European Conference on
  Machine Learning and Principles and Practice of Knowledge Discovery in
  Databases {(ECML} {PKDD} 2018), Dublin, Ireland, September 10th, 2018},
  volume 2284 of {\em {CEUR} Workshop Proceedings}, pages 69--81. CEUR-WS.org,
  2018.

\bibitem[\protect\citeauthoryear{Liu \bgroup \em et al.\egroup
  }{2020}]{DBLP:journals/datamine/LiuLSK20}
Guiliang Liu, Yudong Luo, Oliver Schulte, and Tarak Kharrat.
\newblock Deep soccer analytics: learning an action-value function for
  evaluating soccer players.
\newblock {\em Data Min. Knowl. Discov.}, 34(5):1531--1559, 2020.

\bibitem[\protect\citeauthoryear{Makarov \bgroup \em et al.\egroup
  }{2017}]{makarov:csgo}
Ilya Makarov, Dmitry Savostyanov, Boris Litvyakov, and Dmitry~I. Ignatov.
\newblock Predicting winning team and probabilistic ratings in ''{Dota 2}'' and
  ''{Counter-Strike: Global Offensive}'' video games.
\newblock {\em International Conference on Analysis of Images, Social Networks
  and Texts}, pages 183--196, July 2017.

\bibitem[\protect\citeauthoryear{Song \bgroup \em et al.\egroup
  }{2015}]{song:dota}
Kuangyan Song, Tianyi Zhang, and Chao Ma.
\newblock Predicting the winning side of {Dota 2}.
\newblock Unpublished course project, 2015.

\bibitem[\protect\citeauthoryear{Sun \bgroup \em et al.\egroup
  }{2020}]{DBLP:conf/kdd/SunDSL20}
Xiangyu Sun, Jack Davis, Oliver Schulte, and Guiliang Liu.
\newblock Cracking the black box: Distilling deep sports analytics.
\newblock In Rajesh Gupta, Yan Liu, Jiliang Tang, and B.~Aditya Prakash,
  editors, {\em {KDD} '20: The 26th {ACM} {SIGKDD} Conference on Knowledge
  Discovery and Data Mining, Virtual Event, CA, USA, August 23-27, 2020}, pages
  3154--3162. {ACM}, 2020.

\bibitem[\protect\citeauthoryear{Sutton and Barto}{2018}]{sutton_bartow:rl}
Richard Sutton and Andrew Barto.
\newblock {\em Reinforcement Learning: An Introduction}.
\newblock MIT Press, Cambridge, MA, 2018.

\bibitem[\protect\citeauthoryear{Swaminathan and
  Joachims}{2015}]{swaminathan:sn-iw}
Adith Swaminathan and Thorsten Joachims.
\newblock The self-normalized estimator for counterfactual learning.
\newblock {\em In Advances in neural information processing systems}, pages
  3231--3239, 2015.

\bibitem[\protect\citeauthoryear{Tewari and
  Murphy}{2017}]{tewari:context_bandit}
Ambuj Tewari and Susan~A. Murphy.
\newblock From ads to interventions: Contextual bandits in mobile health.
\newblock In {\em Mobile Health}, pages 495--517. Springer, Cham, 2017.

\bibitem[\protect\citeauthoryear{Williams}{1992}]{williams:rl}
Ronald Williams.
\newblock Simple statistical gradient-following algorithms for connectionist
  reinforcement learning.
\newblock {\em Machine Learning}, 8(3-4):229--256, 1992.

\bibitem[\protect\citeauthoryear{Woodroofe}{1979}]{woodroofe:context_bandit}
Michael Woodroofe.
\newblock A one-armed bandit problem with a concomitant variable.
\newblock {\em Journal of the American Statistical Association},
  74(368):799--806, 1979.

\bibitem[\protect\citeauthoryear{Xenopoulos \bgroup \em et al.\egroup
  }{2020}]{xenopoulos:csgo}
Peter Xenopoulos, Harish Doraiswamy, and Cl{\'{a}}udio~T. Silva.
\newblock Valuing player actions in {Counter-Strike: Global Offensive}.
\newblock In {\em {IEEE} International Conference on Big Data, Big Data 2020,
  Atlanta, GA, USA, December 10-13, 2020}, pages 1283--1292. {IEEE}, 2020.

\bibitem[\protect\citeauthoryear{Yang \bgroup \em et al.\egroup
  }{2016}]{yang:match_pred}
Yifan Yang, Tian Qin, and Yu-Heng Lei.
\newblock Real-time esports match result prediction.
\newblock {\em arXiv preprint}, arXiv:1701.03162, 2016.

\end{thebibliography}

\end{document}